\documentclass[10pt,twocolumn,letterpaper]{article}

\usepackage{cvpr}              %

\usepackage{graphicx}
\usepackage{amsmath}
\usepackage{amssymb}
\usepackage{booktabs}
\usepackage{CJK}
\newcommand{\myparagraph}[1]{\vspace{0.1em}\noindent\textbf{#1}}

\usepackage[pagebackref,breaklinks,colorlinks]{hyperref}

\usepackage[capitalize]{cleveref}
\crefname{section}{Sec.}{Secs.}
\Crefname{section}{Section}{Sections}
\Crefname{table}{Table}{Tables}
\crefname{table}{Tab.}{Tabs.}

\def\cvprPaperID{1201} %
\def\confName{CVPR}
\def\confYear{2022}

\begin{document}

\title{NeuralHOFusion: Neural Volumetric Rendering under Human-object Interactions}

\author{Yuheng Jiang\textsuperscript{1} \;\, Suyi Jiang\textsuperscript{1} \;\, Guoxing Sun\textsuperscript{1} \;\, Zhuo Su\textsuperscript{2} \;\, Kaiwen Guo\textsuperscript{3} \\
\;\, Minye Wu\textsuperscript{4}\;\, Jingyi Yu\textsuperscript{1,5}\;\, Lan Xu\textsuperscript{1,5}} 

\makeatletter
\let\@oldmaketitle\@maketitle%
\renewcommand{\@maketitle}{
	\@oldmaketitle%
	\centering
	\vspace{-8mm}
	{\large \textsuperscript{1}ShanghaiTech University}\quad \quad
	{\large \textsuperscript{2}Tencent}\quad \quad
	{\large \textsuperscript{3}Meta Reasearch Lab}\quad \quad
	{\large \textsuperscript{4}KU leuven}\\
	{\large \textsuperscript{5}Shanghai Engineering Research Center of Intelligent Vision and Imaging}
    
	\vspace{8mm}
}
\makeatother

\maketitle
\pagestyle{empty}
\thispagestyle{empty}

\begin{abstract}
    4D modeling of human-object interactions is critical for numerous applications.
However, efficient volumetric capture and rendering of complex interaction scenarios, especially from sparse inputs, remain challenging.
In this paper, we propose NeuralHOFusion, a neural approach for volumetric human-object capture and rendering using sparse consumer RGBD sensors.
It marries traditional non-rigid fusion with recent neural implicit modeling and blending advances, where the captured humans and objects are layer-wise disentangled.
For geometry modeling, we propose a neural implicit inference scheme with non-rigid key-volume fusion, as well as a template-aid robust object tracking pipeline.
Our scheme enables detailed and complete geometry generation under complex interactions and occlusions.
Moreover, we introduce a layer-wise human-object texture rendering scheme, which combines volumetric and image-based rendering in both spatial and temporal domains to obtain photo-realistic results. 
Extensive experiments demonstrate the effectiveness and efficiency of our approach in synthesizing photo-realistic free-view results under complex human-object interactions.

\end{abstract}

\begin{CJK}{UTF8}{gbsn}
    \section{Introduction}
Human-centric 4D content generation enables numerous applications for VR/AR, telepresence and education.
However, conveniently reconstructing and rendering human activities under human-object interactions remain unsolved.

\begin{figure}[tbp] 
	\centering 
	\includegraphics[width=1\linewidth]{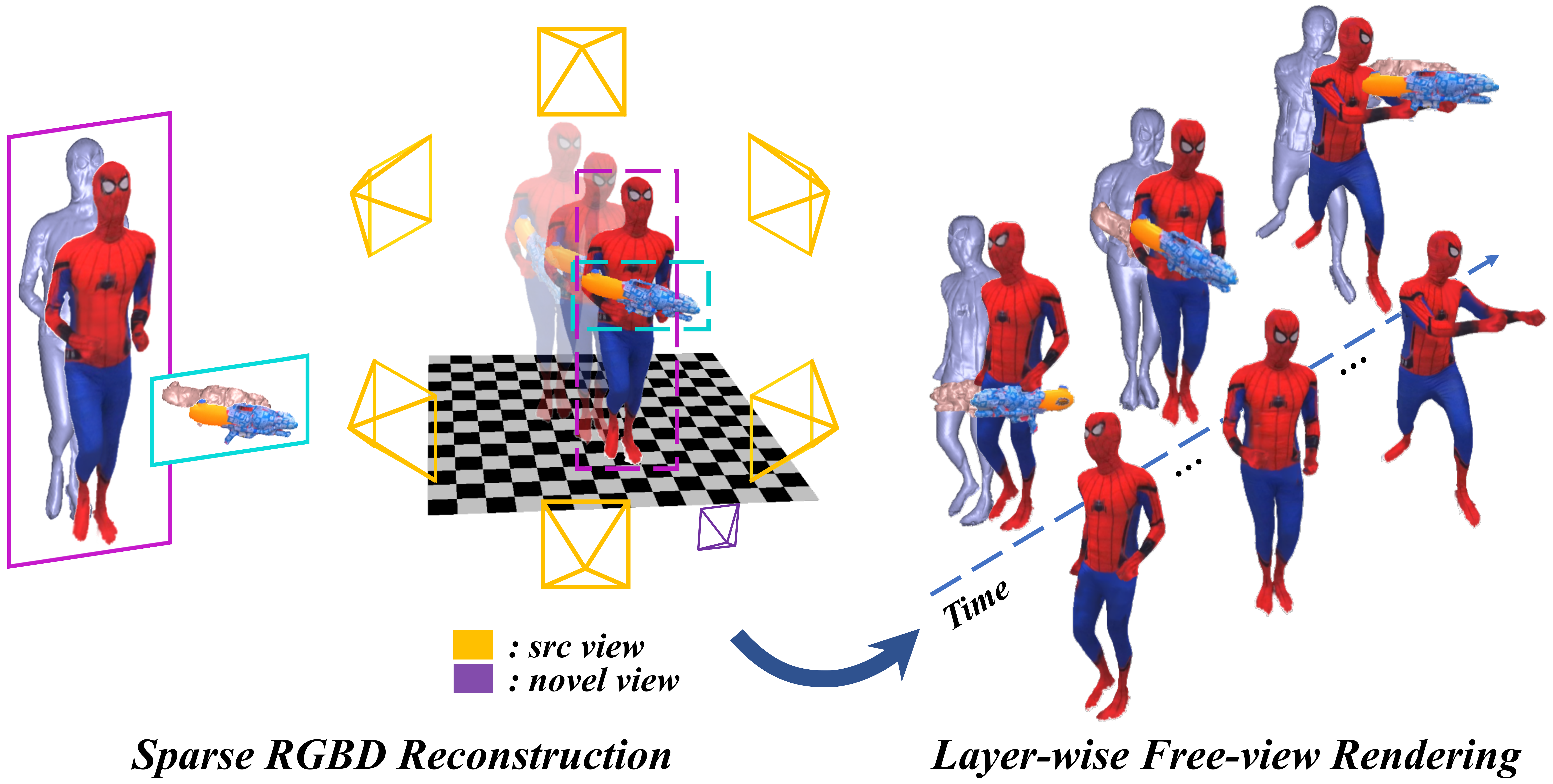} 
	\vspace{-20pt} 
	\caption{Our NeuralHOFusion achieves layer-wise and photo-realistic reconstruction results, using only 6 RGBD cameras.} 
	\label{fig:fig_1_teaser} 
	\vspace{-20pt} 
\end{figure} 

Early high-end solutions~\cite{bradley2008markerless,Gall09b,liu2013markerless,collet2015high,TotalCapture,guo2019relightables} require dense cameras and custom-designed lighting conditions for high-fidelity reconstruction. 
But such a complicated and expensive system setup is undesirable for consumer-level usage.
Light-weight volumetric performance capture is more practical and attractive. 
Early solutions~\cite{li2009robust,HaoliTemplate,Templaterealtime,guo2015robust} rely on pre-scanned templates which are unsuitable for on-the-fly human-object interaction modeling.
The volumetric fusion approaches Fusion4D~\cite{dou2016fusion4d} and Motion2Fusion~\cite{dou2017motion2fusion} further reconstruct complex human-object interaction scenes with topology changes in real-time. 
But they heavily rely on high-quality depth sensors and up to 9 high-end
GPUs, which are infeasible for consumer usage.
Besides, the low-end fusion approaches~\cite{newcombe2015dynamicfusion,KillingFusion2017cvpr,FlyFusion,DoubleFusion,robustfusion} adopt the most handy monocular setup with a temporal fusion pipeline~\cite{KinectFusion}, but suffer from the inherent self-occlusion constraint.
Moreover, the appearance results of the fusion methods are restricted by the limited geometry resolution.

Recent learning-based techniques enable robust human modeling from only light-weight inputs. 
In particular, various approaches~\cite{PIFU_2019ICCV,PIFuHD,suo2021neuralhumanfvv} utilize implicit function to model human geometry, which is also widely adopted in the volumetric capture pipeline~\cite{robustfusion,yu2021function4d,li2021posefusion,Li2020portrait}.
But these methods are restricted to only human without modeling human-object interactions, let alone generating compelling photo-realistic texture.
Similarly, despite the progress for realistic human rendering~\cite{lombardi2019neural,LookinGood,Wu_2020_CVPR,nerf,suo2021neuralhumanfvv}, few researchers explore the neural rendering strategies for human-object interactions, especially under the volumetric capture framework.
On the other hand, various researchers~\cite{PSI2019,zhang2020object,zhang2020phosa,PLACE:3DV:2020,HPS,Hassan:CVPR:2021,PatelCVPR2021,GRAB:2020} model the interactions between humans and the surrounding objects or environments. 
But they only recover the parametric human model rather than reconstructing and rendering the interaction scenes.
Only recently, a few methods~\cite{su2021robustfusion,sun2021HOI-FVV} explicitly model human and object simultaneously in the volumetric capture framework.
But they still cannot handle the interaction scenes, which highly limit the practicality.

In this paper, we present \textit{NeuralHOFusion} -- a neural volumetric human-object capture and rendering system using light-weight consumer RGBD sensors (see Fig.~\ref{fig:fig_1_teaser} for overview).
In stark contrast with existing systems, our approach handles various complex human-object interaction scenarios and even multi-person interactions.
It achieves photo-realistic layer-wise geometry and texture rendering in novel views for both the performers and interacted objects.

Generating such a human-object free-viewpoint video with the layer-wise visual effect whilst maintaining light-weight and efficient setting is non-trivial.
Our key idea is to organically combine traditional volumetric non-rigid fusion pipeline with recent neural implicit modeling and blending advances, besides embracing a layer-wise scene decoupling strategy.
To this end, we first utilize off-the-shelf instance segmentation approach to distinguish the human and object from the six RGBD streams.
For human reconstruction, we propose a fusion-based neural implicit scheme to reason about the human-only geometry details in novel views.
Specifically, it combines pixel-aligned features with an occlusion-aware truncated projective SDF (TSDF) feature~\cite{yu2021function4d}, by utilizing a traditional key-volume non-rigid fusion pipeline~\cite{dou2016fusion4d,FlyFusion} in a human-only manner. 
Such a key-volume fusion-based implicit scheme handles occlusions effectively. 
For object reconstruction, inspired by the recent work~\cite{su2021robustfusion}, we adopt a template-aid robust object tracking pipeline with a specific initialization process for the following neural blending.
Finally, based on the human-object geometry proxy above, we propose a layer-wise neural blending scheme to disentangle human and object for photo-realistic performance rendering.
For the human phase, we combine the image-based rendering with the traditional per-vertex texturing using albedo volume~\cite{UnstructureLan}, through occlusion-aware blending weight learning. 
It enables accurate human appearance rendering in the target view with the level of texture detail in the spatially adjacent input.
For the object rendering, we extend the spatial neural blending into the temporal domain, which learns the blending weight from both the spatial and temporal candidate input views for photo-realistic rendering.
To summarize, our main contributions include:
\begin{itemize} 
	\setlength\itemsep{0em}
	
	\item We present the first neural volumetric capture and rendering system for human-object interaction scenarios using light-weight consumer RGBD sensors.
	
	\item We propose a fusion-based neural implicit inference scheme for detail-preserved human-object reconstruction in an occlusion-aware manner. 
	
	\item We introduce a layer-wise neural rendering scheme, which combines volumetric and image-based rendering in both spatial and temporal domains. 
	
\end{itemize}

    \begin{figure*}[t] 
	\begin{center} 
		\includegraphics[width=\linewidth]{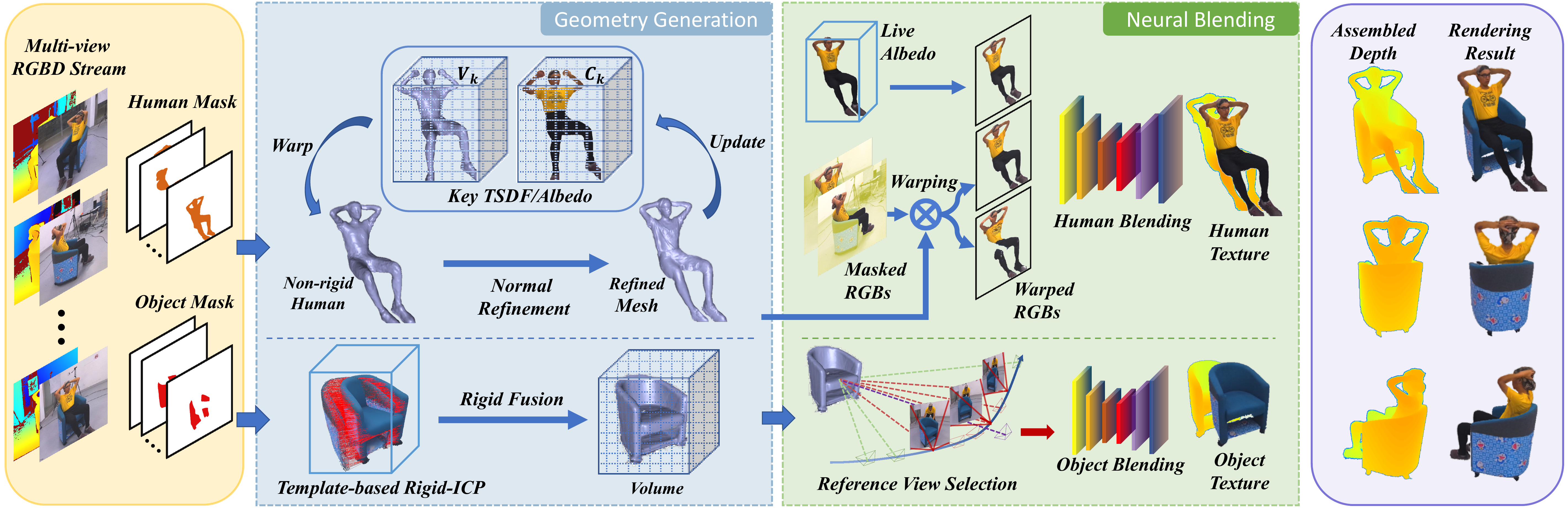} 
	\end{center} 
    \vspace{-20pt}
	\caption{Our approach consists of twos stages. The geometry module includes neural human reconstruction(Sec.~\ref{sec:human geometry}) and  template-aid object fusion (Sec.~\ref{sec:object geometry}), and the blending module includes neural human blending (Sec.~\ref{sec:human texture}) and temporal neural object blending(Sec.~\ref{sec:object texture}).} 
	\label{fig:fig_2_overview} 
	\vspace{-10pt}
\end{figure*}

\section{Related Work} 
\noindent{\textbf{Human-Object Capture.}}
Markerless human-object performance capture techniques have been widely investigated to achieve free-viewpoint video or immersive telepresence.
Early high-end works~\cite{collet2015high,guo2019relightables} use dense cameras for reconstruction and rendering of human and objects through mesh reconstruction and motion tracking, but it is expensive to build the synchronized and calibrated multi-camera systems.
The recent low-end approaches enable light-weight performance capture under the single-RGB setup~\cite{zhang2020phosa,MonoPort,MonoPerfCap}, single-RGBD setup~\cite{su2021robustfusion,burov2021dynamic} or sparse RGBs setup~\cite{suo2021neuralhumanfvv,sun2021HOI-FVV}.
In another line, ~\cite{PSI2019,zhang2020object,PLACE:3DV:2020,HPS,Hassan:CVPR:2021,PatelCVPR2021,GRAB:2020} model the interaction between humans and the objects or the surrounding environments.
PHOSA~\cite{zhang2020phosa} runs human-object capture without any 3D supervision, considering the relationship between human and objects to eliminate ambiguity. But they only recover the naked human template and produce a visually reasonable spatial arrangement.
RobustFusion~\cite{su2021robustfusion} captures human and objects by volumetric fusion, as well as tracks object by Iterative Closest Point. But they cannot handle topology changes and their texture suffers from blur artifacts.
HOI-FVV~\cite{sun2021HOI-FVV} utilizes a decoupling strategy to process human and object respectively under six RGB cameras. Though they show impressive rendering results for human-object interactions, they only process simple pose of human and the inference speed is very slow.
Comparably, our approach achieves high-fidelity capture and rendering for various human-object interactions with complex human pose and severe occlusions at fast speed. 

\noindent{\textbf{Human Volumetric Capture.}}
Volumetric fusion based methods~\cite{newcombe2015dynamicfusion,DoubleFusion,BodyFusion,HybridFusion,FlyCap} allow free-form dynamic reconstruction in a template-free, single-view, real-time way, through updating depth into the canonical model and performing non-rigid deformation.
A series of works are proposed to make volumetric fusion more robust with SIFT features~\cite{innmann2016volumedeform}, human articulated skeleton prior~\cite{DoubleFusion,BodyFusion}, extra IMU sensors~\cite{HybridFusion}, data-driven prior~\cite{robustfusion}, learned correspondences~\cite{bozic2020deepdeform} or neural deformation graph~\cite{bozic2021neural}.
Since these single-view setups suffer from tracking error in the occluded parts, multi-view setups are introduced to mitigate this problem with improved fusion methods.
Fusion4D~\cite{dou2016fusion4d} proposes a key volume updating strategy. Motion2fusion ~\cite{dou2017motion2fusion} incorporates learning-based surface matching into pipeline. UnstructuredFusion~\cite{UnstructureLan} achieves an unstructured multi-view setup. Function4D~\cite{yu2021function4d} combines temporal volumetric fusion and implicit functions to generate complete geometry.
However, these methods either cannot handle modeling human-object interactions or generate photo-realistic rendering results.
Comparably, our approach realizes the abilities of high-fidelity capture and rendering of human-object interactions.

\noindent{\textbf{Neural Rendering and Blending.}}
In the area of photo-realistic novel view synthesis and 3D scene reconstruction, neural rendering shows great power and huge potential.
Various data representations are adopted to obtain better performance and characteristics, such as point-clouds~\cite{Wu_2020_CVPR,aliev2019neural,suo2020neural3d}, voxels~\cite{lombardi2019neural}, texture meshes~\cite{thies2019deferred,liu2019neural} or implicit functions~\cite{park2019deepsdf,nerf} and hybrid neural blending~\cite{suo2021neuralhumanfvv,sun2021HOI-FVV}. 
NHR~\cite{Wu_2020_CVPR} embeds spatial features into sparse dynamic point-clouds, Neural Volumes~\cite{lombardi2019neural} transforms input images into a 3D volume representation by a VAE network.
More recently, \cite{park2020deformable,pumarola2020d,li2020neural,zhao2021humannerf,wang2022fourier} extend neural radiance field~\cite{nerf} into the dynamic setting. 
However, for all approaches above, dense spatial views or full temporal frames are required in training for high fidelity novel view rendering.
Blending based methods learn blending weight for adjacent views and synthesize photo-realistic novel views in a light-weight way. ~\cite{suo2021neuralhumanfvv} uses the occlusion map as guidance for blending weight estimation. ~\cite{sun2021HOI-FVV} incorporates the direction information to reduce artifacts in wide baseline.
However, they cannot handle the occlusion region.
Comparably, our blending with spatial-temporal information, enables recovery of photo-realistic texture of human and objects even under the extreme occlusion region.

\section{Overview}
Given human-object interaction videos under sparse RGBDs setting, NeuralHOFusion can reconstruct high-quality geometries and synthesize layer-wise photo-realistic free-viewpoint videos even in challenging scenarios with extreme poses, occlusions. 
As illustrated in Fig.~\ref{fig:fig_2_overview}, NeuralHOFusion includes two streams for human and objects separately, 
and each stream includes two steps: Geometry Generation and Neural Blending.

\vspace{2mm}\noindent{\bf Geometry Generation.} 
To achieve high-quality human-object geometry for neural rendering under sparse RGBD cameras setting, %
NeuralHOFusion incorporates global temporal information into key volumes.
For humans, we dynamically maintain a key TSDF volume $V_k$ and its albedo volume $C_k$. 
NeuralHOFusion generates a complete geometry of the non-rigid human via a proposed pixel-aligned approach that accompanies either fused TSDF volume $V_t$ of current frame or $V_k$ to assist global reconstruction. 
Besides, normal refinement helps to restore more geometry details. 
We then utilize complete geometry for accurate neural texture blending and key TSDF volume $V_k$ updating. 
For object, we adopt rigid tracking and volumetric fusion to reconstruct the geometry with the aid of the template generated by
an occupancy regression network.

\vspace{2mm}\noindent{\bf Neural Blending.} 
To produce photo-realistic textures based on the above geometries, 
we propose  neural blending schemes to extract features from input textures and predict their blending weights. 
For human, the blending network takes a projected image from albedo volume and adjacent warped RGB images as input, and then predict the blending weight to blend the final texture.
As for object, we reserve non-occluded spatial-temporal observations into an observation-angle group.
We then retrieve ``adjacent views'' from this group to perform temporal blending.
After assembling blended human and object textures, NeuralHOFusion outputs final rendering results. 
    
\section{Method}\label{sec:algorithm} 

\subsection{Neural Human Reconstruction} \label{sec:human geometry}

To reconstruct complete and fine-detailed human geometry, we sequentially perform fusion-based implicit reconstruction and key volume update.%

\begin{figure}[t] 
	\begin{center} 
		\includegraphics[width=1\linewidth]{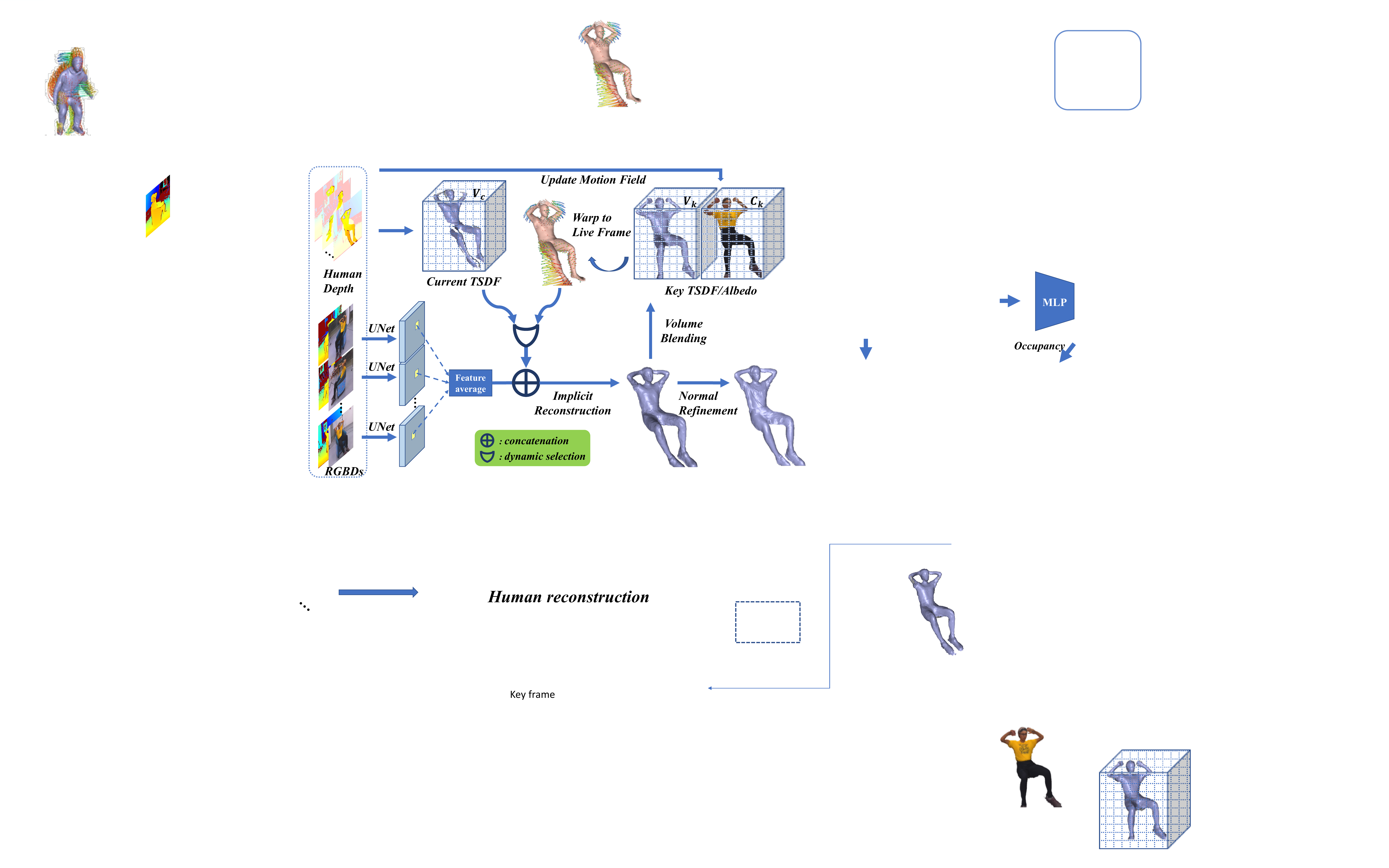}
	\end{center} 
	\vspace{-5mm}
	\caption{Illustration of human geometry reconstruction.(Sec.~\ref{sec:human geometry})} 
	\label{fig:human_geometry_pipeline} 
	\vspace{-10pt}
\end{figure}

\noindent{\bf Fusion-based Implicit Reconstruction.}
Neural networks based on implicit functions are good at complete reconstruction but lack geometry details and temporal consistency, while traditional volumetric capture methods~\cite{newcombe2015dynamicfusion, dou2016fusion4d, UnstructureLan} have enabled the temporal-consistent reconstruction results. From the insight of this complementarity, we follow Function4d~\cite{yu2021function4d} to combine the non-rigid fusion with implicit functions by extracting features from TSDF volume. However, when facing severe occlusions in human-object interactions, ~\cite{yu2021function4d} will fail since their non-rigid fusion in a sliding way could not provide complete TSDF features. Therefore, we non-rigidly fuse a key volume $V_k$ as a reference model, in which we track the motion field from the live frame to it and integrate each depth into this TSDF volume. Note that the motion field is represented by embedded deformation graph(ED-graph)~\cite{sumner2007embedded} and SMPL model~\cite{SMPL:2015}, please refer to~\cite{DoubleFusion,UnstructureLan, robustfusion} for details about this non-rigid fusion process. Then we combine the key volume $V_k$ with current volume $V_t$ which is extracted from current depth and RGBD images to our implicit reconstruction network $f$ for inferring detailed and complete geometry, as shown in Fig.~\ref{fig:human_geometry_pipeline}. We follow the network architecture of PIFu~\cite{PIFU_2019ICCV} to regress an implicit function $f$ to predict occupancy of every 3D point $X$ in the space and formulate it as:
\begin{equation}
\begin{split}
\begin{aligned}
f(\phi(X),\alpha(X), z(X)) &= s : s\in[0.0, 1.0], \\
\phi(X) &= \frac{1}{n}\sum_i^n F_i(\pi_i(X)), \\
\end{aligned}
\end{split}
\label{equ2}
\end{equation}
where $\pi_i(\cdot)$ denotes projection matrix of $i$-th camera;
$z(\cdot)$ is the depth value of $X$. 
$F_i(\pi_i(X))=g(I_i(\pi_i(X)))$ is the image feature of $X$ on RGBD images, $g(\cdot)$ is feature extraction network, $I_i$ is input image.
$\alpha(\cdot)$ represents the queried TSDF value from $V_k$ or $V_t$ of the 3D point $X$, in which $V_t$ gives the detailed geometry information of current frame, while $V_k$ reserves global information in the occluded region. To make the most of them, we introduce a dynamic selection strategy: when the point $X$ is close to the visible human body, we choose $V_t$'s value, $\alpha(X)={V_t}(X)$, otherwise, we choose the value in warped $V_k$.
This operation ensures our method to reconstruct high fidelity geometry in the visible area and reasonable geometry in the occluded part.

\noindent{\bf Key volume update.} 
Our $V_k$ provides a global prior to support the network to infer a complete and temporal-consistent human reconstructions in occlusion scenarios. 
In practice, we reset $V_k$  periodically to reduce misalignment caused by its difference to live frame and enable to handle the topology-changing issues. 
More specifically, we not only fuse the TSDF volume $V_k$ with current depth and the final geometry output through the estimated motion fields~\cite{DoubleFusion} for each frame, we also reset key volume and update ED-graph through re-sampling nodes on the output mesh and SMPL model to re-initialize the motion field at a fixed frequency(40 frames in the paper). 
\begin{figure}[t] 
	\begin{center} 
		\includegraphics[width=1\linewidth]{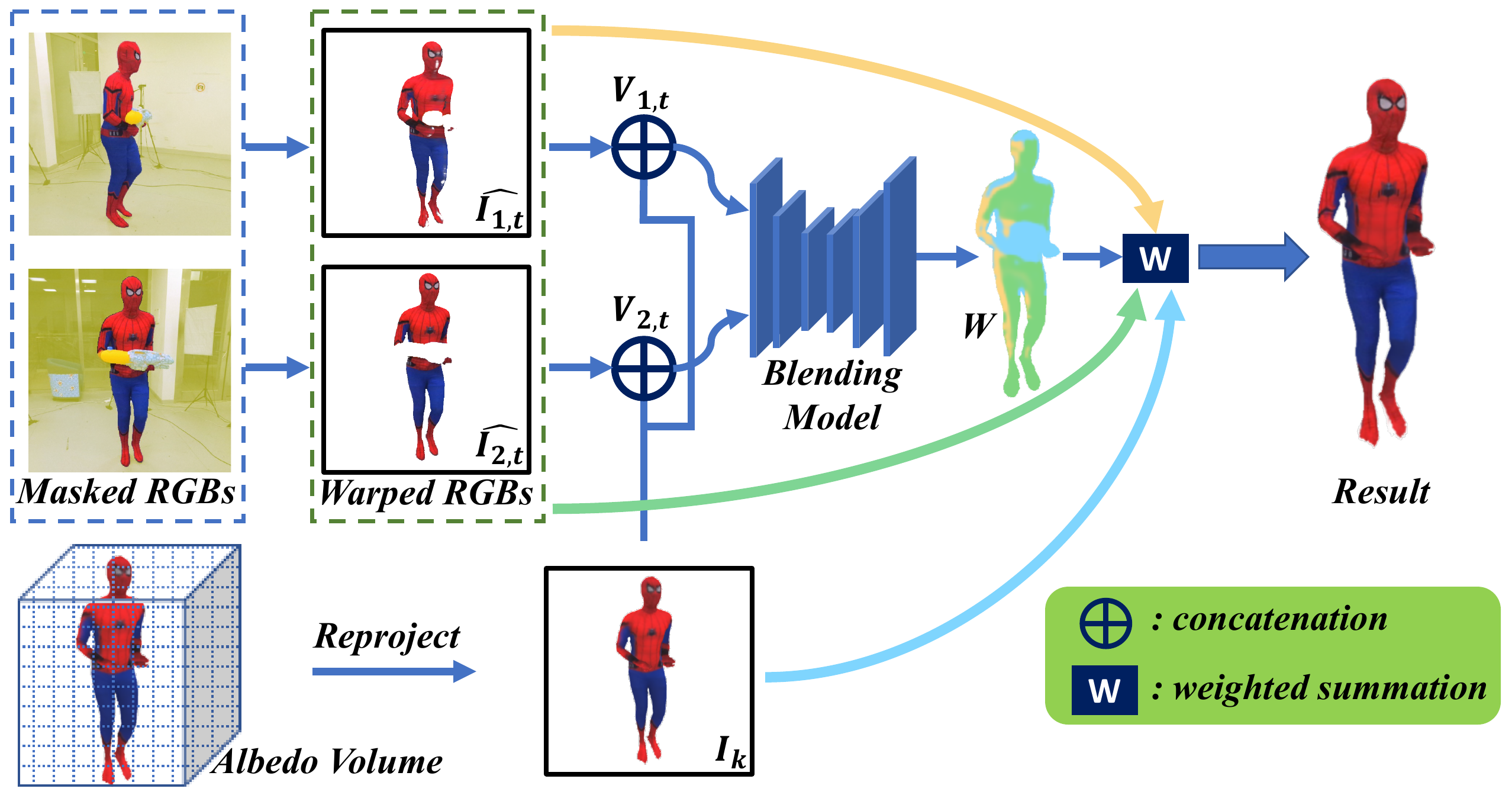}
	\end{center} 
	\vspace{-5mm}
	\caption{Network structure of neural human blending.(Sec.~\ref{sec:human texture})} 
	\label{fig:human_texture_pipeline} 
	\vspace{-15pt}
\end{figure}

\subsection{Template-aid Object Fusion.}   \label{sec:object geometry}
Although naive volumetric fusion~\cite{su2021robustfusion} provides a generally correct object geometry, deteriorated surface caused by the partial depth loss and limited overlap between the matched point-clouds may also occur, as shown in Fig.~\ref{fig:fig_comp_1} (b). 
To achieve stable and accurate object reconstruction, we also incorporate an object template generation module. 
Specifically, this module provides a global cue to regulate object tracking, which greatly improves the robustness of tracking, thus leading to more accurate geometry results.

Firstly, we utilize a data-driven occupancy regression network to generate a complete object template from multi-view RGBD images, in which the formulation resembles Eqn.~\ref{equ2} without TSDF features. 
We then optimize the rigid motions $T$ of the corresponding object point-clouds under the ICP framework as:
\begin{equation}
\begin{split}
\begin{aligned}
\boldsymbol{E}_{\text {object }}(T) = &\lambda_{\text{geo}}\boldsymbol\sum_{(\mathbf{p}, \mathbf{q}) \in \mathcal{R}_{\text{geo}}}\left(\mathbf{n}_{\mathbf{p}}^{T}\left(\mathbf{p}-T \mathbf{q}\right)\right)^{2} +\\
&\lambda_{\text{tem}}\boldsymbol\sum_{(\mathbf{p_t}, \mathbf{q}) \in \mathcal{R}_{\text{tem}}}\left(\mathbf{n}_{\mathbf{p_t}}^{T}\left(\mathbf{p_t}-T \mathbf{q}\right)\right)^{2} +\\
&\lambda_{\text{sp\_o}}\boldsymbol{E}_{\text {sp\_o }},
\end{aligned} 
\end{split}
\end{equation}
where $\mathcal{R}_{\text{geo}}$ is the correspondence pair sets between source point-clouds and fused point-clouds, $\mathcal{R}_{\text{tem}}$ is the correspondence pair sets between source point-clouds and template point-clouds. $\mathbf{q}$ is the point from source, $\mathbf{p}$ is the point from fused model and $\mathbf{p_t}$ from template. $\boldsymbol{E}_{\text {sp\_o }}$ is a term to punish mesh interpenetration, please refer to~\cite{su2021robustfusion} for more details. Finally, with the estimated $T$, source point-clouds are fused into a TSDF volume to update the object geometry.

\subsection{ Neural Human Blending} \label{sec:human texture}

To enable fast novel view synthesis, we adopt a neural blending pipeline to generate photo-realistic and non-occluded human textures under novel view, which incorporates information from the albedo volume maintained in the key frame~\cite{guo2017real}, adjacent input views and local fine-detailed geometry, as illustrated in Fig.~\ref{fig:human_texture_pipeline}. 

\begin{figure}[t] 
	\begin{center} 
		\includegraphics[width=1\linewidth]{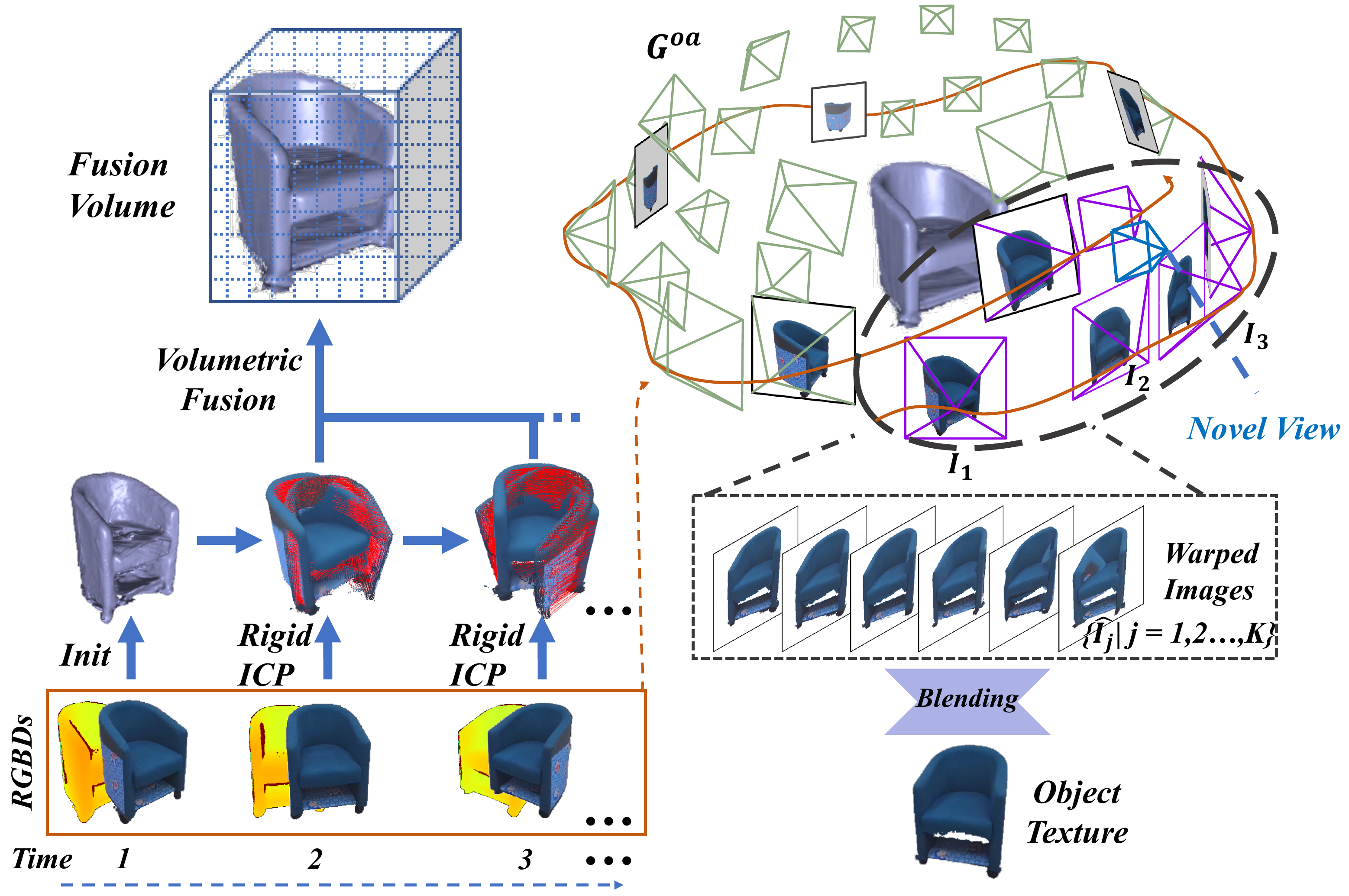}
	\end{center} 
	\vspace{-5mm}
	\caption{Illustration of our object capture and rendering. The left side is the object rigid tracking and fusion(Sec.~\ref{sec:object geometry}). The right side is the temporal neural object blending(~Sec.\ref{sec:object texture}).} 
	\label{fig:object_pipeline} 
	\vspace{-20pt}
\end{figure}

We expand the work ~\cite{suo2021neuralhumanfvv} to define the variance and occlusion maps to learn our blending network. Specifically, we obtain the albedo image and depth map in target view ($I_k$ and $D_k$) using the key albedo volume and occupancy field respectively in \ref{sec:human geometry}. The albedo image $I_k$ lacks details but reserves complete texture. It has rich information for a blending pipeline in the occluded part through providing color candidate and help finding occluded part. We then warp adjacent RGBD maps into target view denoted by $\hat{I_{1,t}}, \hat{I_{2,t}} \hat{D_{1,t}}, \hat{D_{2,t}}$. Subsequently, we calculate the occlusion maps as $O_i = D_k -\hat{D_{i,t}}, i = 1,2$, and compute a per-element variance maps $V_{i,t} = (\hat{I_{i,t}} - I_k)^2,i = 1,2$. Our blending network $\Theta_{HBN}$ utilizes the color information from albedo image and adjacent images, the variance information $V_{i,t}$ and occlusion information $O_i$ to predict pixel-wise blending map $W$, which can be formulated as:
\begin{equation}
\begin{split}
W = \Theta_{HBN}(\hat{I_{1,t}},V_{1,t}, O_1, \hat{I_{2,t}},V_{2,t}, O_2).\\
\end{split}
\end{equation}
Our novel view result of human $I_n$ can be formulated as:
\begin{equation}
\begin{split}
I_n = \hat{W_1}\cdot\hat{I_{1,t}} + \hat{W_2}\cdot\hat{I_{2,t}} + \hat{W_3}\cdot {I_k},
\end{split}
\label{blending_funciton}
\end{equation}
where $\hat{W_{i}}$ denotes blending weights with the sum of 1.0.
\noindent{\bf Normal Refinement.} 
To further improve the quality of geometry in novel view, we follow ~\cite{suo2021neuralhumanfvv} to perform a normal refinement to infer the displacement of the target depth by a normal refinement network $\Theta_{HRN}$. We apply ~\cite{wang2018pix2pixHD} to source view RGBD images $I_{i,t}$, $I_k$ with novel depth ${D_{k}}$, and $I_n$ with novel depth ${D_{k}}$ respectively, we get normal map of source $N_{i,t}$, 
albedo novel view $N_{k}$, and novel view $N_{n}$. We blend adjacent $N_{i,t}$ and $N_k$ with $\hat{W_{i}}$ to get the blended normal map $N_{b}$. $\Theta_{HRN}$ takes $N_{n}$, $N_{b}$ and ${D_{k}}$ as input, and finally predicts the depth displacement.

\begin{figure*}[htbp] 
	\begin{center} 
		\includegraphics[width=1.0\linewidth]{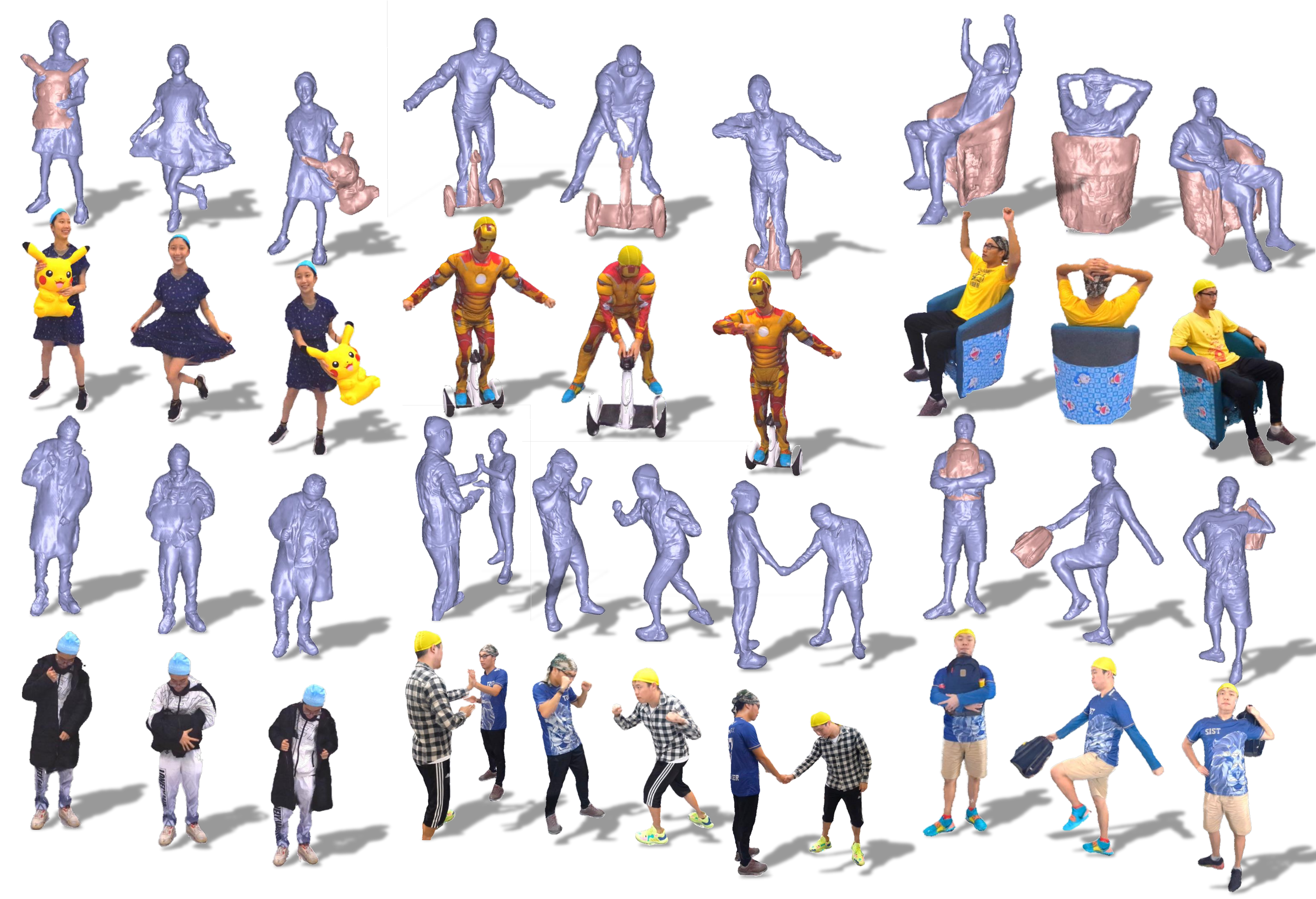} 
	\end{center} 
	\vspace{-20pt}
	\caption{The geometry and texture results of our NeuralHOFusion on various interaction sequences, including
	``floral dress'', ``nesting in a sofa'', ``undressing coat'' and ``shaking hands''.} 
	\label{fig:fig_all}
	\vspace{-20pt}
\end{figure*}

\subsection{Temporal Neural Object Blending.} \label{sec:object texture}

With the temporal observations from object fusion and subsequent tracking, we introduce a strategy to combine these observations into our object blending pipeline smoothly and effectively.
In the initialization stage of object capture, we intentionally show non-occluded images to the cameras and the system works on-line with rigid tracking to collect a group $G^{oa}$ of observation-angle pairs.
In the tracking stage,  non-occluded observation-angle pairs are also added into $G^{oa}$. 
For novel view generation, we interpolate ``nearby views''. More specifically, we follow ~\cite{wang2021ibrnet} to identify a pool of 18 ``nearby views'' from $G^{oa}$ and then randomly sample 6 views from the pool. In this manner, our blending can collect more information in a wider baseline.
We then introduce a temporal neural object blending pipeline to predict novel view object textures as illustrated in Fig.~\ref{fig:object_pipeline}.
This blending network can be formulated as:
\begin{equation}
\begin{split}
W = \Theta_{TBN}(\{\hat{I_{j}},O_j|j=1,2...,6\}).
\end{split}
\end{equation}
$\hat{I_{j}}$ is the warped image of nearby source view $I_j$ from $G^{oa}$. $O_j$ denotes the occlusion map. Novel view images will be generated like Eqn.~\ref{blending_funciton}. 
Although the albedo volume of object can also be used for object blending, we find our blending strategy with $G^{oa}$ observations is sufficient to get good and complete rendering results.

\subsection{Implementation Details} \label{sec:detail}

For human-object segmentation, we first use ~\cite{BGMv2} for background separation then train ~\cite{bolya2020yolact++} to get the initial object coarse mask. Subsequently, We follow ~\cite{su2021robustfusion} to refine the mask. 
For image encoders $g$, we follow ~\cite{suo2021neuralhumanfvv} to use a U-Net, which outputs 64 channels feature maps.
For implicit decoders $f$, we use MLP with skip connections as ~\cite{yu2021function4d}, in which the hidden neurons are (128,128,128,128,128). Furthermore, the loss function of MLP $f$ minimizes the average of mean squared error like ~\cite{PIFU_2019ICCV}.
$\Theta_{HBN}$,$\Theta_{HRN}$ and $\Theta_{TBN}$ adopt the U-Net structure.
For geometry training, we firstly collect 100 human sequences and 40 objects in a dome and utilize existing object meshes from 3D-FUTURE ~\cite{fu20203dfuture}. Then, we place a human in the center of our camera setting and add objects with predefined trajectories to simulate human-object interactions.
We then render RGBD images under our camera parameters and add synthesized noises according to ~\cite{fankhauser2015kinect}. 
For training blending networks, we render the RGBD images, normals and masks under 180 novel views as ground truth. Besides, both L1 loss and perceptual loss~\cite{johnson2016perceptual} are used.
For template-aid object fusion, we use the following empirically determined parameters: $\lambda_{\text{geo}} = 0.8$, $\lambda_{\text{tem}} = 0.2$, $\lambda_{\text{sp\_o}} = 1.0$.

    \section{Experimental Results} 
\vspace{-0.02in}
In this section, we evaluate our NeuralHOFusion in various extremely challenging human-object interactions. 
All the experiments are run on a PC with an Nvidia GeForce RTX3090 GPU and an Intel i7-8700k CPU. 
Capturing six RGBD streams from a synchronized-Kinect system, our method produces high-quality geometry and texture results.
In order to achieve fast performance, we follow ~\cite{yu2021function4d} and ~\cite{suo2021neuralhumanfvv} to implement our entire pipeline on GPU. For each part, the human geometry generation takes 129 ms and 7 GB, the following neural human blending pipeline costs 20 ms and 2.3 GB. The object geometry initialization takes 10 s where the robust object rigid fusion takes around 33 ms. Finally, the neural object blending costs 42 ms. The whole object branch takes around 7 GB of memory consumption.
Various geometry and texture results of our NeuralHOFusion are shown in Fig.~\ref{fig:fig_all}, including different type interactions, and even severe occlusion and topology changes, 
such as nesting in a sofa, shaking hands, and removing clothes. 
\subsection{Comparison} 
\begin{figure*}[htbp] 
	\begin{center} 
		\includegraphics[width=0.95\linewidth]{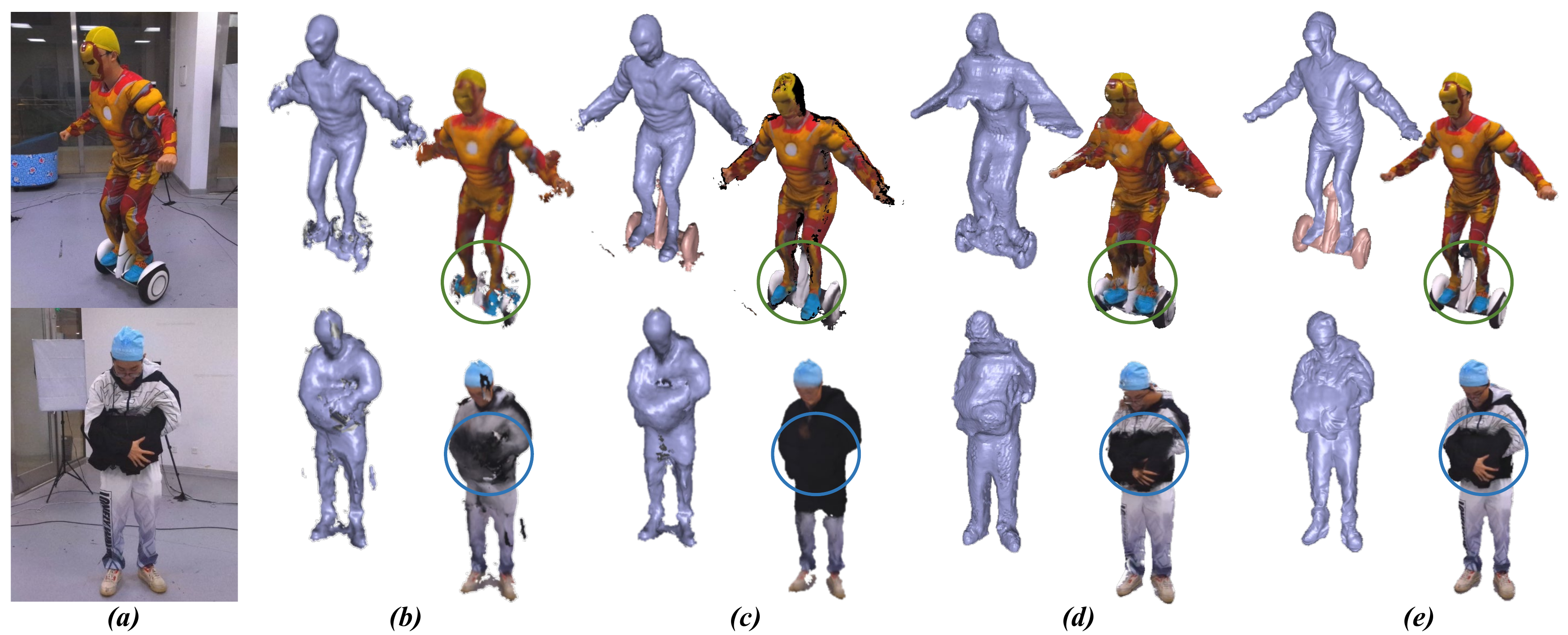} 
	\end{center} 
	\vspace{-20pt}
	\caption{Qualitative comparison. (a) Input images. (b-e) are the geometry and texture results in a novel view from UnstructuredFusion~\cite{UnstructureLan}, RobustFusion~\cite{su2021robustfusion}, NeuralHumanFVV~\cite{suo2021neuralhumanfvv} and ours, respectively.}
	\label{fig:fig_comp_1}
	\vspace{-20pt}
\end{figure*}

We compare our NeuralHOFusion against the start-of-art methods UnstructuredFusion~\cite{UnstructureLan}, RobustFusion~\cite{su2021robustfusion} and NeuralHumanFVV~\cite{suo2021neuralhumanfvv} both in geometry and texture.
As illustrated in Fig.~\ref{fig:fig_comp_1}, UnstructuredFusion~\cite{UnstructureLan} fails to handle the human-object interactions, RobustFusion~\cite{su2021robustfusion} does not support topology changes, and NeuralHumanFVV~\cite{suo2021neuralhumanfvv} cannot reconstruct correct geometry facing different object types. While our NeuralHOFusion achieves more detailed, complete, and isolated geometry results and significantly more photo-realistic rendering results, even under the challenging interactions and extreme human pose.  
Please note that our approach can also enable layer-wise rendering which is not supported by UnstructuredFusion and NeuralHumanFVV.
The quantitative results in Tab.~\ref{table:Comparison_texture} and Tab.~\ref{table:Comparison_geometry} also demonstrate that our approach can achieve consistent better results on all the metrics.

\subsection{Ablation Study} \label{sec:abla} 

\begin{figure}[tbp] 
	\begin{center} 
		\includegraphics[width=0.90\linewidth]{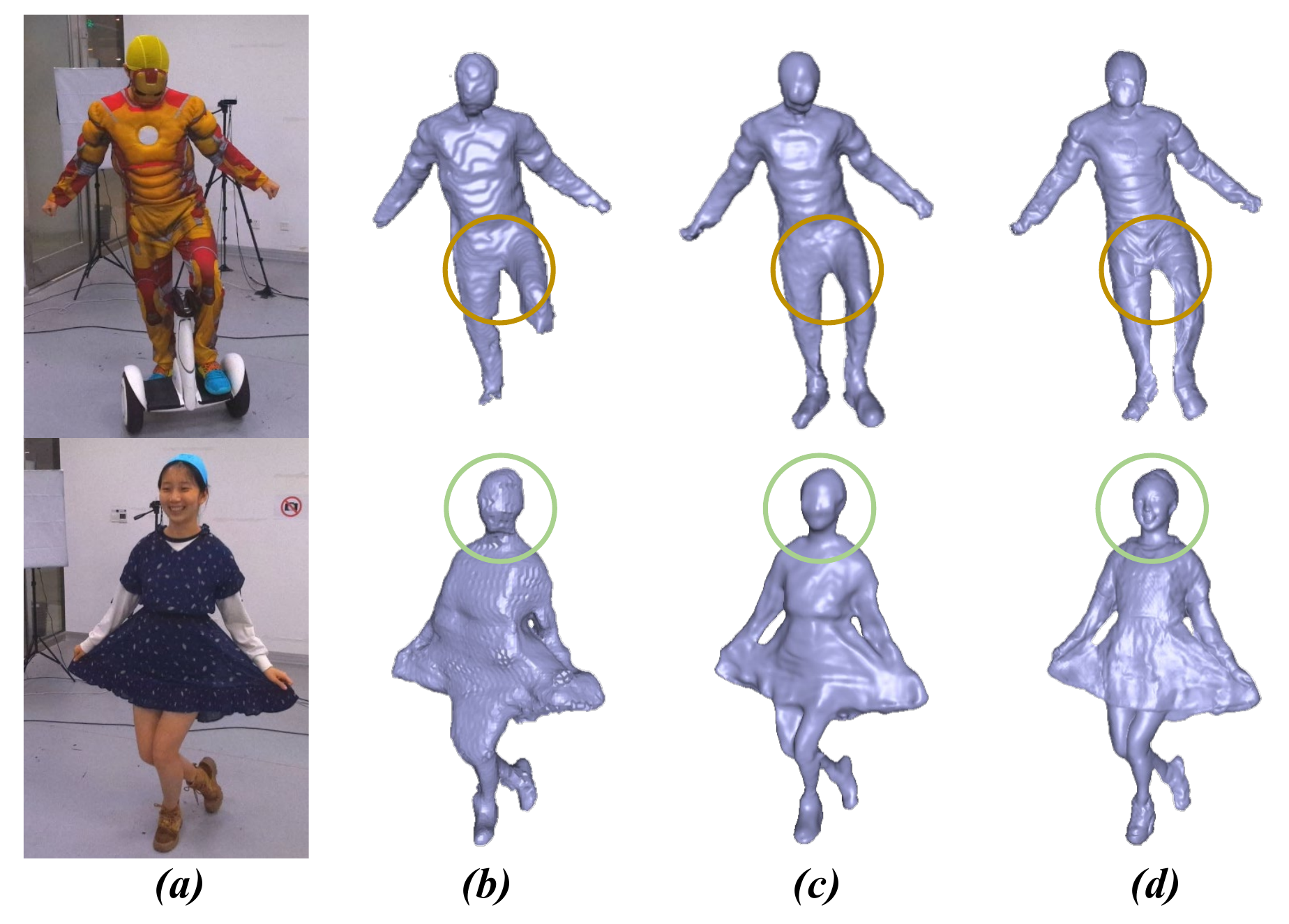} 
	\end{center} 
	\vspace{-20pt}
	\caption{Qualitative evaluation of human geometry generation. (a) Input images. (b) Geometry without TSDF feature; (c) Geometry without normal refinement; (d) Ours.} 
	\label{fig:human_geometry} 
	\vspace{-10pt}
\end{figure}

\myparagraph{Neural Human Geometry Generation.}
As shown in Fig.~\ref{fig:human_geometry} (b), without the TSDF feature, the generation model is unable to recover the occluded human part, and lacks of details,
while after encoding the TSDF feature in Fig.~\ref{fig:human_geometry} (c), the model can generate a complete human body with mid-level geometry details such as the clothing wrinkles but still suffers from over-smooth results, especially on the facial area. 
In contrast, our full approach with normal refinement achieves detailed human geometry reconstruction as shown in Fig.~\ref{fig:human_geometry} (d). 
For further quantitative analysis, we evaluate each component using \textbf{CD} and \textbf{P2S} as shown in Tab.~\ref{table:eval_ho_geometry}, which highlights the contributions of our each component.

\begin{figure}[tbp] 
	\begin{center} 
		\includegraphics[width=0.90\linewidth]{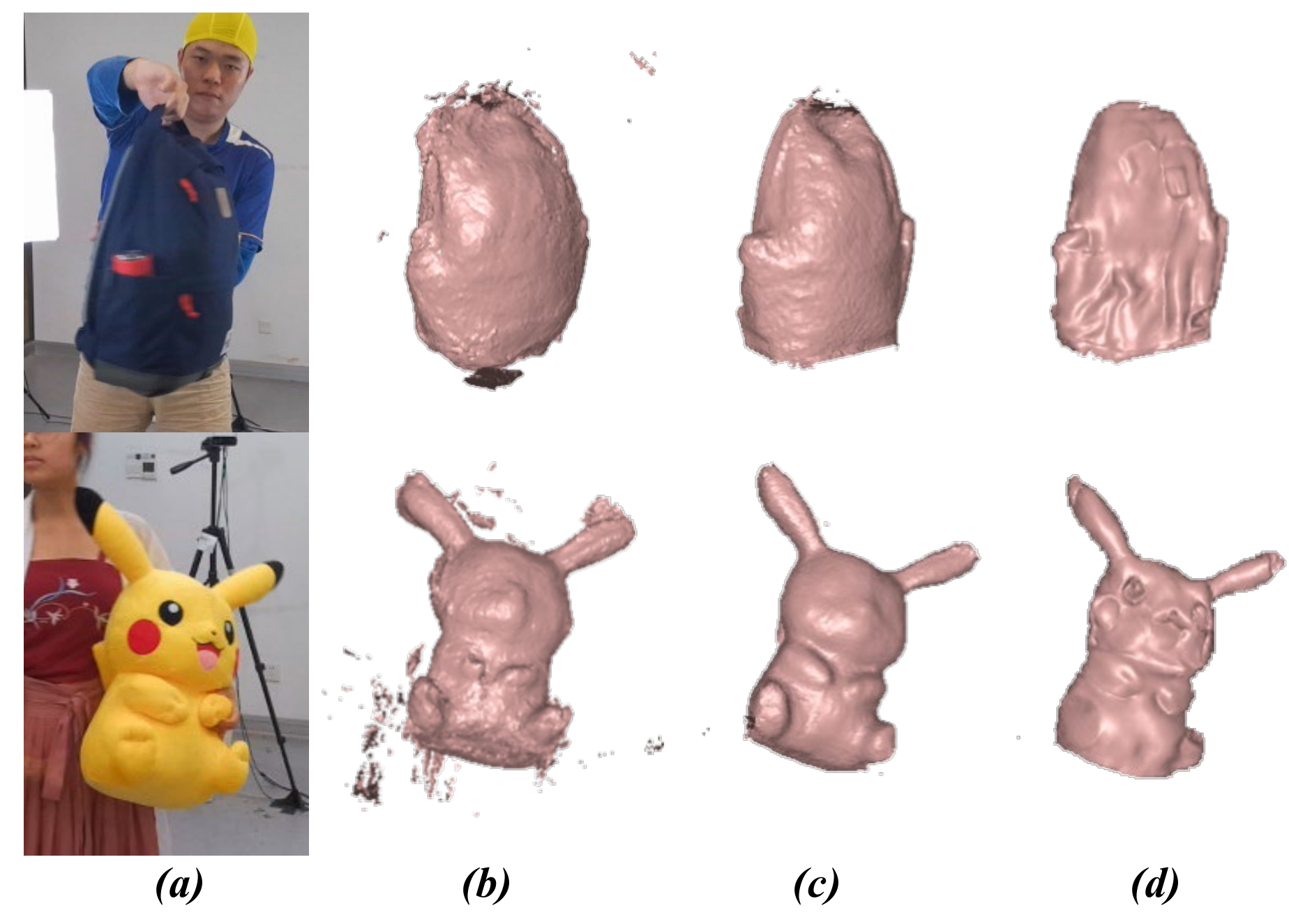} 
	\end{center} 
	\vspace{-20pt}
	\caption{Qualitative evaluation of object geometry generation. (a) Input images. (b) Geometry without template; (c) Geometry without normal refinement; (d) Ours.} 
	\label{fig:object_geometry} 
	\vspace{-10pt}
\end{figure}

\myparagraph{Template-aid Object Fusion.}
As shown in Fig.~\ref{fig:object_geometry} (b), without the template produced by MLP, simple rigid ICP is prone to slight mismatches, which affects the generation of fused mesh.
Furthermore, due to the limitation of volume resolution, template-aid rigid ICP still suffers from the over-smooth issues as shown in Fig.~\ref{fig:object_geometry} (c). 
In contrast, our full pipeline with refinement can recover more detailed object geometry. 
Further quantitative analysis in Tab.~\ref{table:eval_ho_geometry} demonstrates that our method achieves higher accuracy.  

\begin{figure}[t] 
	\begin{center} 
		\includegraphics[width=1.0\linewidth]{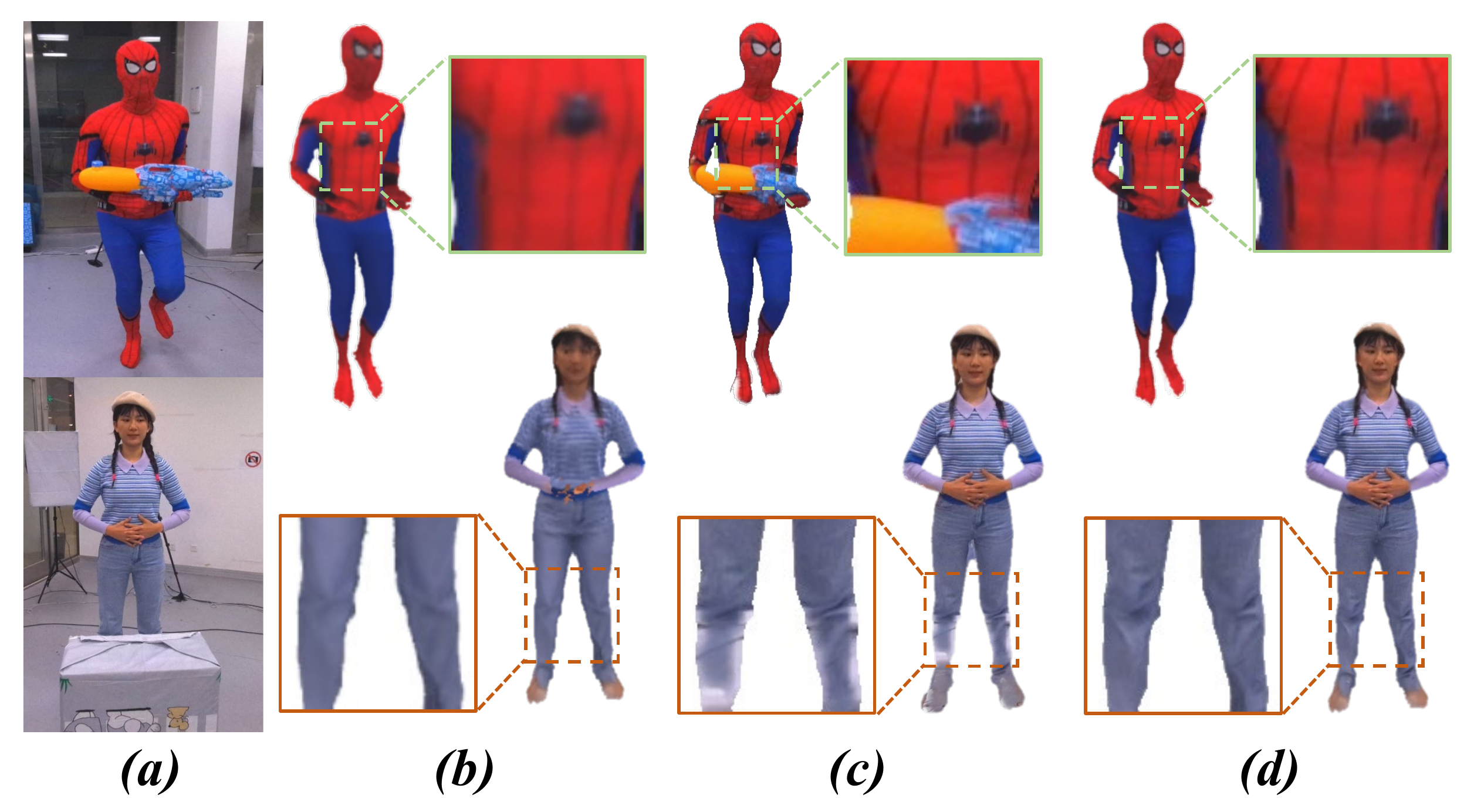} 
	\end{center} 
	\vspace{-20pt}
	\caption{Qualitative evaluation of neural blending. (a) Input image; (b) Per-vertex texture; (c) Naive neural blending; (d) Ours.} 
	\label{fig:human_texture} 
	\vspace{-20pt}
\end{figure}

\myparagraph{Neural Human Blending.}
\begin{table}[t]
	\begin{center}
		\centering
		\caption{Quantitative comparison of rendering results.}
		\vspace{-10pt}
		\label{table:Comparison_texture}
		\resizebox{0.45\textwidth}{!}{
			\begin{tabular}{l|cccc}
				\hline
				Method      & PSNR$\uparrow$ & SSIM$\uparrow$ & MAE $\downarrow$  \\
				\hline
				UnstructuredFusion ~\cite{UnstructureLan}\qquad\qquad & 22.456            & 0.937         & 3.058    \\
				RobustFusion  ~\cite{su2021robustfusion}       & 26.537         & 0.941         & 1.868  \\
				NeuralhumanFVV  ~\cite{suo2021neuralhumanfvv} &      27.526  & 0.979 & 1.131 \\
				Ours        & \textbf{33.59}   & \textbf{0.984} & \textbf{0.627} \\
				\hline
			\end{tabular}
		}
		\vspace{-20pt}
	\end{center}
\end{table}
\begin{table}[t]
	\begin{center}
		\centering
		\caption{Quantitative comparison of geometry reconstructions.}
		\vspace{-10pt}
		\label{table:Comparison_geometry}
		\resizebox{0.45\textwidth}{!}{
			\begin{tabular}{l|cccc}
				\hline
				Method      & P2S$\times$10$^{-4}$$\downarrow$ 
				& Chamfer$\times$10$^{-4}$$\downarrow$ \\
				\hline
				Multi-PIFu \cite{PIFU_2019ICCV}     &  \qquad\qquad\qquad14.475\qquad\qquad\qquad            & 10.564             \\
				RobustFusion  \cite{robustfusion}      & \qquad\qquad\qquad5.770\qquad\qquad\qquad            & 6.375               \\
				Ours        & \qquad\qquad\qquad\textbf{2.692}\qquad\qquad\qquad   & \textbf{2.853}  \\
				\hline
			\end{tabular}
		}
		\vspace{-20pt}
	\end{center}
\end{table}
In Fig.~\ref{fig:human_texture}, we evaluate different variants of texturing schemes with the same geometry proxy. 
The texture extracted from albedo volume as shown in Fig.~\ref{fig:human_texture} (b) is blurred, 
while the naive neural blending results in Fig.~\ref{fig:human_texture} (c) suffer from severe block artifacts, which blends the object texture to the human. 
In contrast, our full neural human blending scheme achieves both photo-realistic and complete texture results as shown in  Fig.~\ref{fig:human_texture} (d). Besides, we also make a comparison on a synthetic sequence with 400 frames and generate 180 different target views to evaluate. The Tab.~\ref{table:eval_ho_texture} demonstrates that our method achieves higher accuracy.

\begin{table}[t]
    \Huge
	\begin{center}
		\centering
		\caption{Quantitative evaluation of reconstruction schemes.}
		\vspace{-10pt}
		\label{table:eval_ho_geometry}
		\resizebox{0.45\textwidth}{!}{
			\begin{tabular}{l|cccc}
				\hline
				Method      & P2S$\times$10$^{-4}$  $\downarrow$ 
				& Chamfer  $\times$  10  $^{-4}$  $\downarrow$   \\
				\hline
				w/o TSDF (human)      & 7.7407            & 7.969         \\
				w/o normal refinement (human)      & 3.1425     & 3.4086       \\
				neural human geometry generation        & \textbf{2.855}   & \textbf{3.239}  \\
				\hline
				\hline
				w/o template (object)      & 35.138            & 19.383            \\
				w/o normal refinement (object)      & 11.529     & 9.256         \\
				template-aid object fusion        & \textbf{11.480}   & \textbf{9.166}  \\
				\hline
			\end{tabular}
		}
		\vspace{-20pt}
	\end{center}
\end{table}

\begin{table}[t]
    \footnotesize
	\begin{center}
		\centering
		\caption{Quantitative evaluation of texturing schemes.}
		\vspace{-10pt}
		\label{table:eval_ho_texture}
		\resizebox{0.45\textwidth}{!}{
			\begin{tabular}{l|cccc}
				\hline
				Method      & PSNR$\uparrow$ & SSIM$\uparrow$ & MAE $\downarrow$  \\
				\hline
				albedo volume (human)       & 26.758            & 0.925          & 2.167   \\
				naive neural blending (human)         & 25.983         & 0.962         & 1.735 \\
				neural human blending        & \textbf{30.040}   & \textbf{0.968} & \textbf{0.945} \\
				\hline
				\hline
				albedo volume (object)       & 33.455            & 0.950          & 0.564   \\
				naive neural blending (object)          & 30.760         & 0.968         & 1.255 \\
				temporal neural object blending        & \textbf{37.901}   & \textbf{0.971} & \textbf{0.376} \\
				\hline
			\end{tabular}
		}
		\vspace{-20pt}
	\end{center}
\end{table}

\begin{figure}[t] 
	\begin{center} 
		\includegraphics[width=0.95\linewidth]{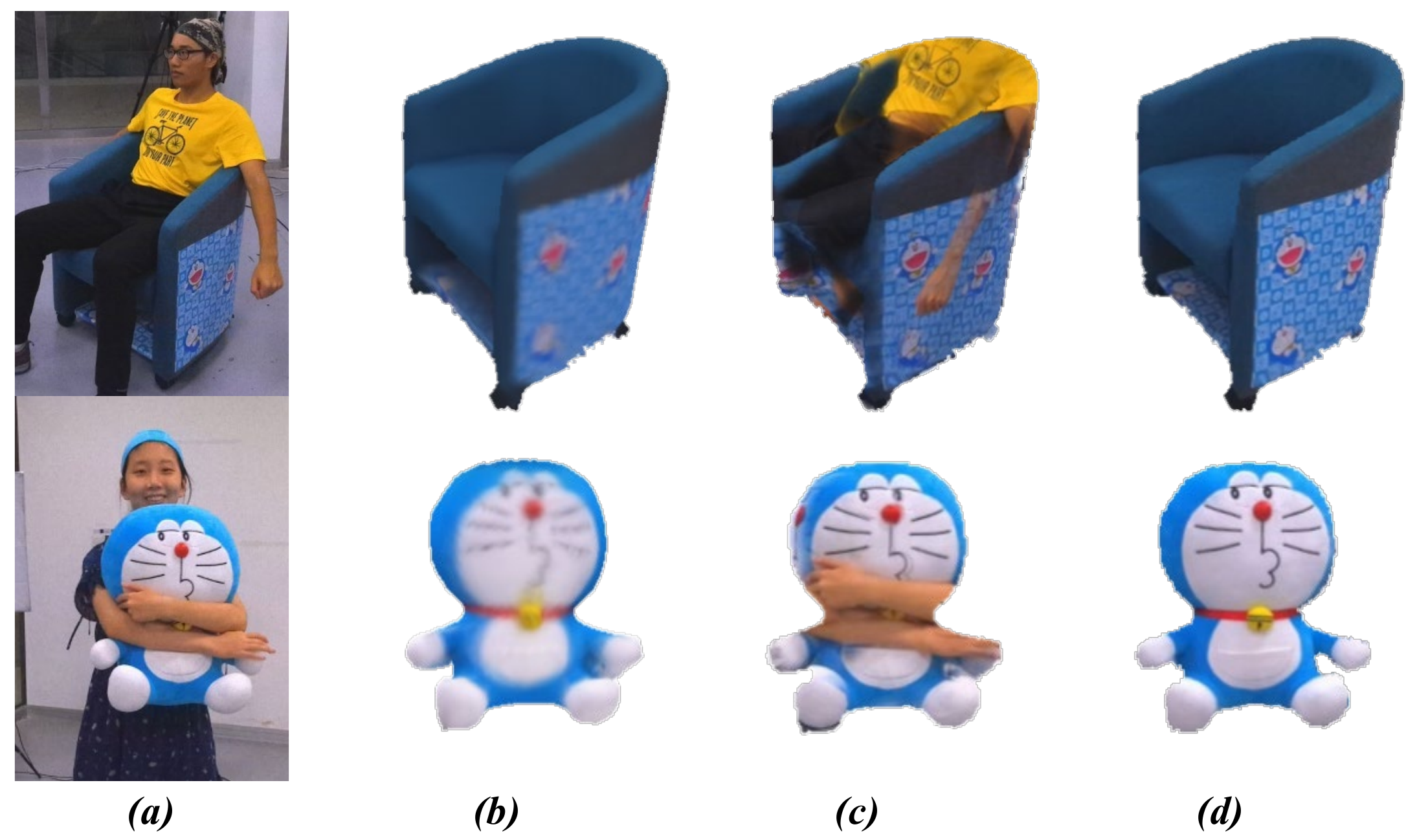} 
	\end{center} 
	\vspace{-20pt}
	\caption{Qualitative evaluation of neural blending scheme on objects. (a) Input images; (b) Per-vertex texture; (c) Naive neural texture blending; (d) Ours.} 
	\label{fig:object_texture} 
	\vspace{-10pt}
\end{figure}

\myparagraph{Temporal Neural Object Blending.}
As for evaluation of object texturing, Fig.~\ref{fig:object_texture} (b) demonstrates that texture fusion scheme leads to blur,
Fig.~\ref{fig:object_texture} (c) shows that the naive neural texturing blending scheme wrongly recovers the texture which belongs to the human part.  
In contrast, our temporal neural object blending makes full use of both the previous non-occluded frames and the current frame.  Therefore, we can faithfully recover the accurate texture even when some parts are sereve occluded in Fig.~\ref{fig:object_texture} (d). The quantitative experiments on synthetic object sequences can refer to Tab.~\ref{table:eval_ho_texture}.

\begin{figure}[tbp] 
	\begin{center} 
		\includegraphics[width=0.95\linewidth]{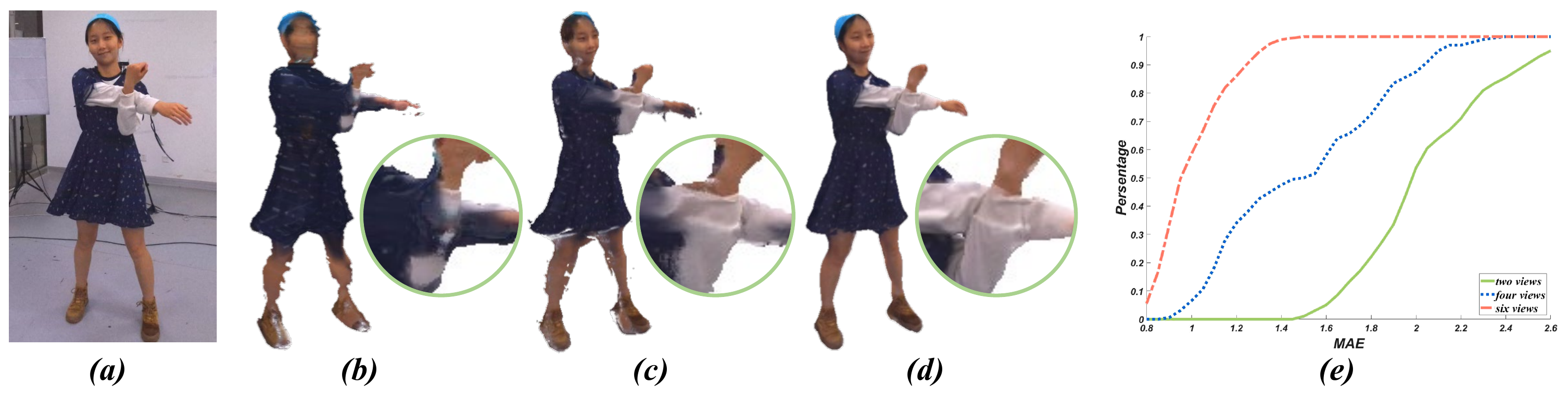} 
	\end{center} 
	\vspace{-20pt}
	\caption{Evaluation of the number input camera views. (a) Reference image. (b, c, d) Our reconstructed texture results using two, four and six cameras, respectively. (e) Cumulative distribution function of the mean absolute error.} 
	\label{fig:fig_camera_number} 
	\vspace{-4mm}
\end{figure}

\myparagraph{\noindent{\bf Camera Number.}}
We evaluate the influence of input view number in our multi-view setting,
where the cameras are placed around a circle uniformly and numbered from 0 to 5.
We compare the results of two-camera system (0, 3), four-camera system (0, 1, 3, 4) and six-camera system.
As shown in Fig.~\ref{fig:fig_camera_number}, without sufficient camera views, the reconstructed geometry is a little downgrading, while the textured results significantly get worse. 
Empirically, we find the system with six-camera produces good results in a compromise of camera number and quality.

\subsection{Limitation}

Although NeuralHOFusion can perform detailed and complete reconstruction and layer-wise photo-realistic rendering under complex human-object interactions by fully utilizing temporal observations, it still owns some limitations.
First, our method heavily relies on instance segmentation method, bad segmentation will lead to awful geometry and interlaced texture.
Besides, since depth sensors lack observations from specific materials such as shaggy hair and yarn clothes, our method cannot get good geometry to these areas. 
A semantic-aware implicit function on different human parts will be critical for such problem.
Furthermore, our approach will produce texture-copy artifacts after normal refinement.
Our current pipeline models human and objects separately, and it is an interesting direction to build a physical framework such as ~\cite{shimada2020physcap}.

\section {Conclusion} 
We have presented a practical neural volumetric capture and rendering approach for complex human-object interaction scenes, using sparse  RGBD cameras.
By combining traditional non-rigid fusion with neural implicit modeling and blending, our system achieves detailed and realistic results with the unique layer-wise viewing experience.
Our fusion-based neural implicit inference and template-aid object tracking enable detailed and complete geometry generation under occlusions, while our texturing scheme combines volumetric and image-based rendering in both spatial and temporal domains to synthesize photo-realistic texture. 
Our experimental results demonstrate the effectiveness of NeuralHOFusion in complex interaction scenarios with various poses and clothing types.
We believe that our approach is a critical step to virtually but realistic teleport human performances under complex interactions, with many potential applications like consumer-level telepresence, active object scanning and human behavior analysis.  \\     
\noindent{\bf Acknowledgements.} This work was supported by Shanghai YangFan Program (21YF1429500), Shanghai Local college capacity building program (22010502800).

\end{CJK}

{\small
\bibliographystyle{ieee_fullname}
\bibliography{reference}
}

\end{document}